%% file: main.tex
\documentclass[preprint, 10pt, twocolumn, nonatbib]{elsarticle}
\makeatletter\let\c@author\relax\makeatother
\usepackage[backend=biber, sorting=none, style=ieee]{biblatex}
\usepackage{graphicx, caption, subcaption, multirow} 
\usepackage{amsfonts,amssymb,amsmath,amsthm}	
\usepackage{booktabs,multirow,tabularx}

\usepackage{placeins}[section]

\usepackage[T1]{fontenc}
\usepackage{lmodern}
\usepackage{hyperref}

\usepackage{tikz}
\usepackage{float}
\usepackage{flushend}

\hypersetup{
    colorlinks=true,
    linkcolor=blue,
    citecolor=blue,
    filecolor=magenta,
    urlcolor=blue,
    pdftitle={Multistep and Multivariate Time Series Explainability with GradientSHAP: A Wind Power Case Study},
    pdfpagemode=FullScreen,
}

\usepackage{multicol}

\journal{Computational Intelligence}
\begin{document}
\begin{onecolumn}
\begin{frontmatter}

\title{Multistep and Multivariate Time Series Explainability with GradientSHAP: A Wind Power Case Study}

\clearpage
\address[1]{Department of Engineering, University of Exeter, Exeter, EX4 4QF, United Kingdom}
\address[2]{Department of Computer Science, University of Exeter, Exeter, EX4 4QF, United Kingdom}

\author[1]{Robbie Southam\corref{cor1}}
\ead{rs1117@exeter.ac.uk}
\author[1]{Zhong Fan}
\ead{z.fan@exeter.ac.uk}
\author[2]{Ke Li}
\ead{k.li@exeter.ac.uk}
\cortext[cor1]{Corresponding author: Robbie Southam}

\begin{abstract}
    As the effects of climate change become increasingly apparent, the demand for renewable energy sources such as wind power continues to grow. However, renewables' variable and intermittent nature presents significant challenges in their integration into energy grids. Artificial intelligence (AI) has emerged as a powerful tool to address these issues. Still, its adoption in the energy sector has been hindered by the opacity of many AI models, which often function as black boxes. This paper introduces a carefully adapted GradientSHAP approach to provide interpretability for multivariate, multistep wind power forecasting — a largely underexplored area. This capability is critical for day-ahead grid management and, to the best of our knowledge, this is the first study to apply GradientSHAP in the context of wind power forecasting. To address the inherent dimensionality, we developed novel visualizations that deliver both local and global explanations, uncovering intricate relationships between environmental factors, site-specific conditions, and temporal dynamics. While GradientSHAP is not an out-of-the-box solution, our findings demonstrate that, with thoughtful handling of SHAP values and domain expertise, it can offer valuable insights into model behaviour, supporting more informed and trustworthy decision-making. Nevertheless, its complexity poses challenges for practical implementation. Future research should focus on streamlining these interpretability methods to make them more actionable for routine deployment in critical energy applications.
\end{abstract}
\begin{graphicalabstract}
\includegraphics{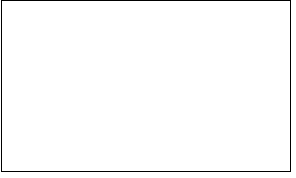}
\end{graphicalabstract}

\begin{highlights}
\item We present a novel application of GradientSHAP to day-ahead wind power forecasting, addressing a critical gap in explainable AI for renewable energy systems.
\item We provide a practical guide for deriving meaningful insights from SHAP methods, combining domain knowledge with innovative visualizations tailored for complex time series data.
\item Our results reveal critical temporal, spatial, and feature-specific dynamics in wind power forecasting, validating known relationships and uncovering novel interactions through GradientSHAP analysis.
\item We advance the limited body of work on multivariate, multitarget time series explainability, which is crucial for improving renewable energy forecasting and operational transparency.
\end{highlights}


\end{frontmatter}
\end{onecolumn}

\twocolumn
\input{Sections/Introduction}

\input{Sections/RelatedWork}
\input{Sections/Methodology}
\input{Sections/ResultsDiscussion}

\input{Sections/Conclusion}


\appendix
\input{Sections/Appendix}

\printbibliography
\end{document}

%% file: Sections/Introduction.tex
\section{Introduction}
Achieving global net-zero goals necessitates urgent, transformative efforts to reduce carbon emissions and accelerate the shift to renewable energy sources (RESs), prompting a fundamental reshaping of the global energy framework \cite{haronADViCEAIDecarbonisation2023}. Central to this transition is the rapid deployment of RESs such as wind\cite{haronADViCEAIDecarbonisation2023, dellannaGreenJobsEnergy2021, przychodzenDeterminantsRenewableEnergy2020}. Wind power, as one of the fastest-growing RESs, has emerged as a critical driver in reducing greenhouse gas (GHG) emissions, while also presenting substantial economic benefits \cite{zhangWindTunnelExperiment2021, sivaramakrishnaiahNumericalExperimentalInvestigations2024}. However, the variability and intermittency of renewable energy pose significant challenges for integration, impacting grid stability and operational reliability \cite{przychodzenDeterminantsRenewableEnergy2020, nguyenPotentialExplainableArtificial2024,haronADViCEAIDecarbonisation2023}. 

Accurate short-term wind power forecasting is indispensable for addressing these challenges. It plays a vital role in managing grid load effectively, ensuring reliable energy dispatch, and optimising penetration and connection rate  \cite{machlevExplainableArtificialIntelligence2022, ersozShortReviewExplainable2022, liShorttermWindPower2022, tsaiReviewModernWind2023}. However, despite substantial advancements in forecasting accuracy, top-performing models often function as ``black boxes,'' with their decision processes obscured. This lack of transparency hinders stakeholder trust and limits the adoption of AI in high-stakes domains such as energy\cite{alsaighAIExplainabilityGovernance2023}. 

Explainable Artificial Intelligence (XAI) approaches, particularly post-hoc feature attribution methods such as Local Interpretable Model-agnostic Explanations (LIME) and SHapley Additive exPlanations (SHAP)\cite{lundbergUnifiedApproachInterpreting2017}, have gained traction as tools to illuminate model logic and improve interpretability \cite{lundbergUnifiedApproachInterpreting2017, machlevExplainableArtificialIntelligence2022, ersozShortReviewExplainable2022, nguyenPotentialExplainableArtificial2024}. These methods enable grid operators and policymakers to glean insights into model behaviour by highlighting the contribution of each feature to model outputs, enhancing their capacity to manage grid stability and develop informed policies\cite{operatorProbabilisticMachineLearning2022, xuReviewInterpretableMachine2022, nguyenPotentialExplainableArtificial2024}.


SHAP, grounded in cooperative game theory, apportions feature importance based on each feature’s marginal contribution to model outputs. This approach satisfies key principles of local accuracy, missingness, and consistency, ensuring fair and reliable attribution \cite{lundbergUnifiedApproachInterpreting2017}. However, traditional SHAP implementations, such as DeepSHAP and KernelSHAP, are limited by computational demands or strict assumptions like linearity or feature independence \cite{AAS2021103502, janzingFeatureRelevanceQuantification2020}. GradientSHAP, an extension specifically designed for differentiable models, addresses these constraints by leveraging expected gradients (EG) \cite{erionImprovingPerformanceDeep2020} to provide more efficient and reliable explanations while maintaining the same desirable principles as SHAP\cite{erionImprovingPerformanceDeep2020}.

To the best of our knowledge, this paper represents the first application of GradientSHAP in wind power forecasting and extends the very limited body of work on multivariate, multistep time series explainability. Through a comprehensive analysis of GradientSHAP values, supported by several novel visualizations, we uncover known physical relationships and explore intriguing feature interactions. Additionally, the model's predictions exhibited a strong reliance on typical temporal patterns, including diurnal and, to a lesser extent, seasonal trends. 

While GradientSHAP, combined with domain expertise, demonstrated considerable potential in enhancing interpretability and fostering transparency, its reliance on expert knowledge limits its broader applicability. 

Our analysis, which largely exposes the model's decision process, reveals significant alignment with real-world dynamics in wind power forecasting. This demonstrates the utility of GradientSHAP, when combined with domain knowledge, for enhancing interpretability in short-term forecasting models. By improving transparency, this approach empowers grid operators and policymakers to make more informed decisions, contributing to greater grid stability and operational efficiency. Ultimately, by fostering trust in AI systems, this work supports the broader adoption of explainable AI in the energy sector.

The remainder of this paper is organised as follows. \ref{sec:related_work} reviews relevant literature on feature attribution in energy forecasting, with a particular focus on SHAP and its applications in multivariate, multistep forecasting tasks. Section \ref{sec:methodology} outlines the GradientSHAP methodology and describes the feature engineering and model architecture used in our forecasting approach. Section \ref{sec:results_discussion} presents the results, starting with global and local SHAP analyses and examining feature interactions, followed by a detailed examination of temporal patterns and the seasonal dynamics of feature importance. Finally, Section \ref{sec:conclusion} concludes with a summary of findings and directions for future work.

%% file: Sections/RelatedWork.tex
\section{Related Work}
\label{sec:related_work}
Despite SHAP’s prominence as a feature attribution method in energy forecasting, most applications remain focused on demand-side tasks\cite{gugliermettiFutureDirectionMachine2024, bialekDatasetLevelExplanation2022, cuiEnergyConsumptionPrediction2024, maarifEnergyUsageForecasting2023}, with notably fewer studies on supply-side forecasting\cite{zhangComparisonExplanationForecasting2021, dolgintsevaComparisonMultistepForecasting2024, vanzylHarnessingEXplainableArtificial2024}. Additionally, these studies largely employ Recurrent Neural Network (RNN)-based models, ensemble methods, and Artificial Neural Networks (ANNs) with limited innovation in the SHAP methodology itself. Significantly, few address the complexity of multivariate or multistep timeseries tasks\cite{dolgintsevaComparisonMultistepForecasting2024, cakirogluDatadrivenInterpretableEnsemble2024, vanzylHarnessingEXplainableArtificial2024} and even fewer address both\cite{bialekDatasetLevelExplanation2022}, underscoring the need for further exploration and novel approaches\cite{simicXAIMethodsNeural2021}.

For example, on the demand side, \citeauthor{vanzylHarnessingEXplainableArtificial2024} evaluated the efficacy of Grad-CAM and SHAP for feature selection in a multistep load forecasting scenario using a Convolutional Neural Network (CNN), generating explanations with DeepSHAP\cite{vanzylHarnessingEXplainableArtificial2024}. They found both methods to enhance forecasting accuracy and model efficiency by identifying and excluding irrelevant features. Although Grad-CAM was computationally more efficient, SHAP better-captured features that degraded forecast accuracy, though it inaccurately ranked feature importance, likely due to DeepSHAP's assumptions. The authors highlight the need for refinement in XAI methods within time-series forecasting to improve interpretability.

To the best of our knowledge, only \citeauthor{bialekDatasetLevelExplanation2022} tackles the problem of both multivariate and multistep time series by applying SHAP for feature attribution in a heat demand forecasting model for the Warsaw District Heating Network\cite{bialekDatasetLevelExplanation2022}. Their use of DeepSHAP aimed to provide both local and global explanations. Due to the high degree of correlation among features, the authors aggregated SHAP values to capture meaningful groups, such as aggregating temperature-related features to observe their collective influence on heat demand. To interpret temporal dependencies, they employed visualisation methods such as SHAP attribution versus \textit{cos\_yday} for seasonality, aggregated feature group plots, 3D scatter plots for delayed temporal features, and box plots to illustrate distributions across values of temporal features. While this study demonstrates that SHAP can offer valuable insights for multistep multivariate forecasting, the authors note that interpreting the results still requires significant domain expertise. Moreover, the analysis relied heavily on engineered features without state-of-the-art models, limiting its generalisability and applicability to more advanced model architectures. Importantly, the independence assumption inherent in DeepSHAP further restricts the accuracy of the attributions, given the high correlation between many features.

In the context of energy forecasting \citeauthor{maarifEnergyUsageForecasting2023} investigates energy consumption predictions using LSTM networks and SHAP explainability\cite{maarifEnergyUsageForecasting2023}. The authors developed single-layer, double-layer, and bidirectional LSTM models, and tested them on a steel industry dataset, finding that the double-layer model offered the best performance. SHAP summary plots were used for interpretability, specifically layered violin and bar plots, to display the contribution of each feature, ranked in descending order. The bar chart presents the mean absolute SHAP values for each feature, while the layered violin plot provides a more detailed view by showing the distribution of SHAP values for each feature along with corresponding feature values. This analysis highlighted the significant influence of leading current reactive power and time of day, with minimal impact from features like \textit{CO\_2} emissions and weekly status. 

While GradientSHAP appears highly suitable for the context of this paper and forecasting problems as a whole (see Section \ref{sec:feature_attribution}), its application within time-series forecasting, particularly in the energy domain, remains limited. There are only a handful of studies in time series contexts \cite{cossuDriftingExplanationsContinual2024,
vanzylHarnessingEXplainableArtificial2024, turbeEvaluationPosthocInterpretability2023, nguyenRobustExplainerRecommendation2024}. Although some studies indicate that GradientSHAP outperforms alternative methods \cite{serramazzaEvaluatingExplanationMethods2023, najjarFeatureAttributionMethods2023},  to the best of our knowledge, it is limited to a single paper within the energy domain\cite{vanzylHarnessingEXplainableArtificial2024}. Furthermore, many studies employing GradientSHAP omit the number of background samples used, which complicates comparisons between methods.

While \cite{schlegelRigorousEvaluationXAI2019} present perturbation and sequence analysis techniques to benchmark feature attribution methods in univariate timeseries forecasting\cite{schlegelRigorousEvaluationXAI2019}, \citeauthor{veerappaValidationXAIExplanations2022a} adapt this to multivariate classification and incorporates GradientSHAP into the benchmark\cite{veerappaValidationXAIExplanations2022a}. Specifically, they evaluate GradientSHAP, DeepSHAP, path-integrated gradients (PIG) and a time-series adapted version of LIME called LimeforTime in the maritime domain. Additionally, the authors extend the prior evaluation methodology with a technique called ``noise'' to assess the quality of explanations for both perturbation and sequence analysis; investigating how much the model's output is affected if noise is added to the input. The authors deem these methods suitable for multivariate time series, showing that DeepSHAP and GradientSHAP work robustly in all cases; while being one of the few papers to indicate the number of background samples ($5000$). While the performance of DeepSHAP and GradientSHAP were essentially the same, GradientSHAP has significantly less execution time than all other methods and $\sim 13\%$ that of DeepSHAP. Further, they suggest the use of temporal heatmaps for more intuitive explanations.

This paper seeks to address these gaps by applying GradientSHAP in a novel multivariate, multistep energy forecasting context, an area that remains largely unexplored. By leveraging GradientSHAP’s computational efficiency, and it's ability to operate with reduced assumptions compared to traditional methods, we aim to enhance both the interpretability and operational viability of modern energy forecasting models, as well as provide a tangible example of utilising GradientSHAP for domain experts. 

%% file: Sections/Methodology.tex
\section{Methodology}
\label{sec:methodology}
\subsection{Dataset}
\label{sec:dataset}

We use WIND\cite{draxl2015wind} by The National Renewable Energy Laboratory (NREL), providing seven years of meteorological data in $5$ minute granularity from more than $126,000$ sites.

\subsubsection{Feature Selection}
To ensure varied environmental conditions, $4$ states have been selected spread out around the US: Washington, New York, Iowa and Texas. Three sites are sampled for each, giving a total of $12$ sites, containing five years. The resulting data were split into the train, validation and test sets temporally. Here, the years $2008$-$2010$, $2011$ and $2012$ were used for each. Additionally, we reduce the granularity of the dataset to an hour by averaging.

To determine the number of lags we use the Autocorrelation Function (ACF), Partial Autocorrelation Function (PACF), and Cross-Correlation Function (CCF)\cite{fengDatadrivenMultimodelMethodology2017, fengDatadrivenMultimodelMethodology2017} on the training set and determine that a lag of $48$ is best for balancing the model learning and insights we can obtain from the explanation. We also test for stationarity using Augmented Dickey-Fuller (ADF) and Kwiatkowski-Phillips-Schmidt-Shin (KPSS) tests before this, to ensure interpretations of correlation are meaningful.

Similarly, we average the dataset by month, day, year, and hour to observe any seasonality. We find that the most prevalent trends are both hourly and monthly.

\subsubsection{Preprocessing}
To ensure the model was able to make the most of the data, we performed additional encoding on some features. To ensure that daily and seasonal patterns could be captured better, we used cyclical encoding on the \textit{day} and \textit{month} features. Similarly, we performed the same on \textit{wind direction} to retain the circular structure and prevent discontinuity between $0\deg$ and $360\deg$. Finally, since we only use $11$ sites from such a big pool, we integer encode the site ID.

To generate time-series samples for training the model, we utilise the rolling window approach. At each time step $t$, a window $\mathcal{W}_t$ was created that included $48$ historical time steps for input and $24$ time steps for prediction. An example of this can be seen in Figure \ref{fig:rollingwindow}.


\subsection{Model Architectures}
Recurrent neural networks (RNNs) and their variants, particularly long short-term memory (LSTM) networks, are widely used for multivariate time series forecasting, especially for multistep problems \cite{chandelReviewDeepLearning2023}. LSTMs are favoured due to their ability to capture dependencies across longer sequences by mitigating the vanishing gradient problem that affects traditional RNNs. 

To develop an effective forecasting model, we explore several architectures, beginning with a baseline LSTM model and progressing to more intricate encoder-decoder (seq2seq) structures. Two RNN-based networks, LSTMs in this case, make up the encoder/decoder, or sequence to sequence (Seq2Seq) architecture\cite{sutskeverSequenceSequenceLearning2014}. The encoder first processes and compresses the input sequence into a fixed-sized context vector, which the decoder then uses to iteratively produce each step of the output sequence. However, the fixed context vector can limit the model's ability to retain detailed information over long input sequences.

To enhance the model’s predictive accuracy and focus on relevant input segments during decoding, we integrate \textit{Luong attention} \cite{luongEffectiveApproachesAttentionbased2015} into the Seq2Seq architecture. Luong attention allows the model to dynamically adjust the importance of each input time step in generating each output prediction. By actively comparing the current decoder hidden state, $\mathbf{h}_t$, with each encoder output, $\mathbf{h}_s$, Luong attention assigns weights (alignment scores) that emphasise the most relevant portions of the input sequence. This approach enables a focused alignment on the recent $48$-hour context, which is particularly useful in day-ahead forecasting where immediate past conditions are strong predictors of near-future outcomes.

All three of Luong’s alignment scoring functions are considered in our model \cite{luongEffectiveApproachesAttentionbased2015}:
\begin{equation}
    \text{score}(\mathbf{h}_t, \mathbf{h}_s) =
    \begin{cases}
        \mathbf{h}_t^\top \mathbf{h}_s & \text{dot} \\ 
        \mathbf{h}_t^\top \mathbf{W}_a \mathbf{h}_s & \text{general} \\
        \mathbf{v}_a^\top \tanh{(\mathbf{W}_a [\mathbf{h}_t; \mathbf{h}_s])} & \text{concat} \\
    \end{cases}
\end{equation}
Each method provides different levels of flexibility in learning relationships within the data:
\begin{itemize}
    \item \textit{Dot}: Computationally efficient, directly computes the dot product between $\mathbf{h}_t$ and $\mathbf{h}_s$.
    \item \textit{General}: Uses a learnable weight matrix $\mathbf{W}_a$ to scale $\mathbf{h}_s$, capturing adaptable contributions across input steps.
    \item \textit{Concat}: Concatenates $\mathbf{h}_t$ with $\mathbf{h}_s$ and applies a non-linear transformation, offering higher flexibility but at greater computational cost.
\end{itemize}

While self-attention and Transformer-based cross-attention mechanisms excel at capturing long-range dependencies, our temporal encoding of the \textit{day} and \textit{month} features should serve as a proxy for these longer-term dependencies. Luong attention on the other hand, focuses on short-term dependencies within recent observations, which are particularly relevant for day-ahead predictions and is less computationally expensive. Further, using explicit alignment scores enhance interpretability and enabling stakeholders to trace forecasted outputs back to specific recent observations.

Finally, an embedding layer processes the \textit{site ID} feature to capture location-specific factors, such as microclimates or terrain effects, which can vary between sites and influence wind 

Finally, an embedding layer processes the site ID feature to capture location-specific factors, such as microclimates or terrain effects, which can vary between sites and influence wind and power patterns.

\subsection{Optimisation}
To achieve more accurate and logically sound explanations and ensure further reliability in generated forecasts, we optimise both hyperparameters and the choice of architecture. Given the dimensionality of the search space and the computational expense of training, the models we employ Bayesian Optimisation (BO) to optimise the choice of architecture and remaining parameters. Additionally, we use early stopping and a learning rate scheduler to ensure that models do not overfit and reduce training time. Fixed hyperparameters and ones that were found to work well overall can be seen in Table \ref{tab:hps}.

Generally, the encoder-decoder architecture performed the best with attention. The \textit{general} scoring function within the attention mechanism performed the best, compared to \textit{concat} and \textit{dot}. A detailed optimisation summary can be found in Table \ref{tab:hp_optimization} and the final architecture in Figure \ref{fig:summary}.

\subsection{Explainability by Feature Attribution}
\label{sec:feature_attribution}

Feature attribution has become a core post-hoc interpretability method across numerous fields, including energy\cite{zhangComparisonExplanationForecasting2021, vanzylHarnessingEXplainableArtificial2024}, driven by the increasing complexity of machine learning models.  Feature attribution attempts to address the black-box challenge by quantifying the importance of each input feature for a given prediction, thus helping domain experts to gain insights into model decisions \cite{doshivelez2017rigorousscienceinterpretablemachine}. 

SHAP was originally introduced as a unified framework that combines multiple approaches to additive feature importance, such as LIME, DeepLIFT, and Layer-wise Relevance Propagation (LRP), under a single theoretical foundation based on cooperative game theory\cite{lundbergUnifiedApproachInterpreting2017}. In SHAP, each feature is treated as a \textit{player} in a game. Players work together to make the model prediction which is viewed as the \textit{payout} and must be fairly distributed. The marginal contribution of each feature to the prediction is calculated by averaging its impact across all possible combinations of features, or \textit{coalitions}. This yields the Shapley value for each feature $i$:
\begin{equation}
\begin{gathered}
    \phi_{i, SHAP} = \sum_{S \subseteq F \setminus \{i\}} \frac{|S|! (|F| - |S| - 1)!}{|F|!} \\
    \Big[ f_{S \cup \{i\}}(x_{S \cup \{i\}}) - f_S(x_S) \Big]
\end{gathered}
\end{equation} where $F$ is the set of all features and $S$ corresponds to a subset of features.

SHAP ensures several desirable properties that make it well-suited for model interpretability:

\begin{itemize}
    \item \textbf{Local Accuracy}: The SHAP value sum approximates the model’s output at a given instance.
    \item \textbf{Missingness}: Features not in the original input are assigned zero importance.
    \item \textbf{Consistency}: If a feature's contribution increases in the model, its Shapley value does not decrease.
\end{itemize}

Given these properties, \citeauthor{lundbergUnifiedApproachInterpreting2017} prove that methods not based on shapely violate  \textit{local accuracy} and/or \textit{consistency}.

GradientSHAP, is a more recent, gradient-based extension of SHAP specifically for differentiable models and is a synthesis of ideas from Integrated Gradients (IG)\cite{sundararajan2017axiomatic} and SmoothGrad.  IG is a path attribution method corresponding to Aumann-Shapley values\cite{aumannValuesNonatomicGames1974}; a cost-sharing method that extends Shapley values to infinite-player games. GradientSHAP follows the same approach as IG but is enhanced by sampling across multiple baselines and averaging gradients along paths between each baseline and the input. Many traditional SHAP variants are based on the observational conditional expectation, which leads to importance attributed to irrelevant features, and misleading explanations when features are correlated\cite{janzingFeatureRelevanceQuantification2020, sundararajanManyShapleyValues2020}. By integrating over a background set, GradientSHAP follows a more interventional approach and produces explanations that are ``true to the model'' rather than ``true to the data''\cite{chenTrueModelTrue2020}. 

In contrast to SHAP, GradientSHAP extends the original IG calculation, calculating Shapley values by taking the expectation of gradients over interpolated points from each baseline to the target input:

\begin{equation}
\begin{gathered}
    \phi_{i, EG}(f, x; D) = \mathbb{E}_{x' \sim D, \alpha \sim U(0, 1)} \\
    \Big[(x_i - x'_i) \times \frac{\partial f(x' + \alpha \times (x - x'))}{\partial x_i} \Big]
\end{gathered}
\end{equation}

where $x_\prime$ is a sample from the original data distribution $D$, and $x \prime$ is a \textit{reference} sample. Similarly, $\alpha$ is a scaling factor, and $\frac{\delta F}{\delta x_i}$ is the gradient with respect to feature $i$.

Not only does EG satisfy all axioms proposed by \citeauthor{lundbergUnifiedApproachInterpreting2017} (which are similar to the properties mentioned above) for SHAP, several newer axioms are also satisfied \cite{sundararajan2017axiomatic}:
\begin{itemize}
    \item \textit{Sensitivity(a)}: For every input and baseline that differ in one feature but have different predictions then the differing feature should be given a non-zero attribution.
    \item \textit{Sensitivity(b)} - \textit{Dummy}: This axiom is satisfied by any interventional. It requires that a feature will get zero attribution if it does not affect the model's output.
    \item \textit{Implementation Invariance}: The attribution for any two, functionally equivalent networks should always be identical.
\end{itemize}

Adhering to these axioms is ideal for complex multistep and multivariate time series problems, where many feature interactions and dependencies can exist. By leveraging EG, GradientSHAP provides a computationally efficient\cite{veerappaValidationXAIExplanations2022a} and theoretically grounded approach to day-ahead forecasting; enabling a thorough exploration of lags, features, prediction steps and their interactions.

\subsubsection{Utilising GradientSHAP}
To ensure a balanced trade-off between computational efficiency and explanation fidelity, we followed the guidance\cite{veerappaValidationXAIExplanations2022a} and set the background sample size to $5000$. As indicated in \cite{yuanEmpiricalStudyEffect2022}, the number of background samples can significantly impact GradientSHAP’s stability, where too few samples may introduce noise and too many can incur unnecessary computational costs. For testing and explanation generation, we used $300$ samples, calculating SHAP values for each feature across all $48$ historical time steps and $24$ future predictions.

A notable challenge encountered during implementation was the incompatibility between GradientSHAP and embeddings, which prevented SHAP from attributing any contribution to the \textit{site ID} feature due to its integer encoding. This issue, documented in the SHAP package, arises because integer-encoded embeddings lack gradients, resulting in no attribution to these features in the output. To address this, we implemented forward and backward hooks to sum the gradients of the \textit{site ID} embedding vector, ensuring proper gradient flow and enabling accurate feature attribution. This solution preserved the integrity of the explanations and allowed GradientSHAP to provide a complete view of feature contributions, crucial for reliable decision-making in wind power forecasting.

\subsection{Experimental Setup}
Model training, inference and explanation generation have been performed on a system with a Nvidia GeForce GTX 1080 Ti, Intel Core i7-14700k and 48GB of RAM. PyTorch\cite{paszkePyTorchImperativeStyle2019} has been used for model development, training and inference using the GPU.

%% file: Sections/ResultsDiscussion.tex
\section{Results \& Discussion}
\label{sec:results_discussion}
This section provides a comprehensive breakdown of the model's feature importance through various plots, forming global and local feature explanations. We start with summary plots and progressively uncover interaction effects and temporal dynamics.

\subsection{Summary Plots}
\label{sec:local}
\begin{figure*}[htb]
    \centering
    \begin{subfigure}[t]{.495\textwidth}
        \centering\includegraphics[width=1\textwidth]{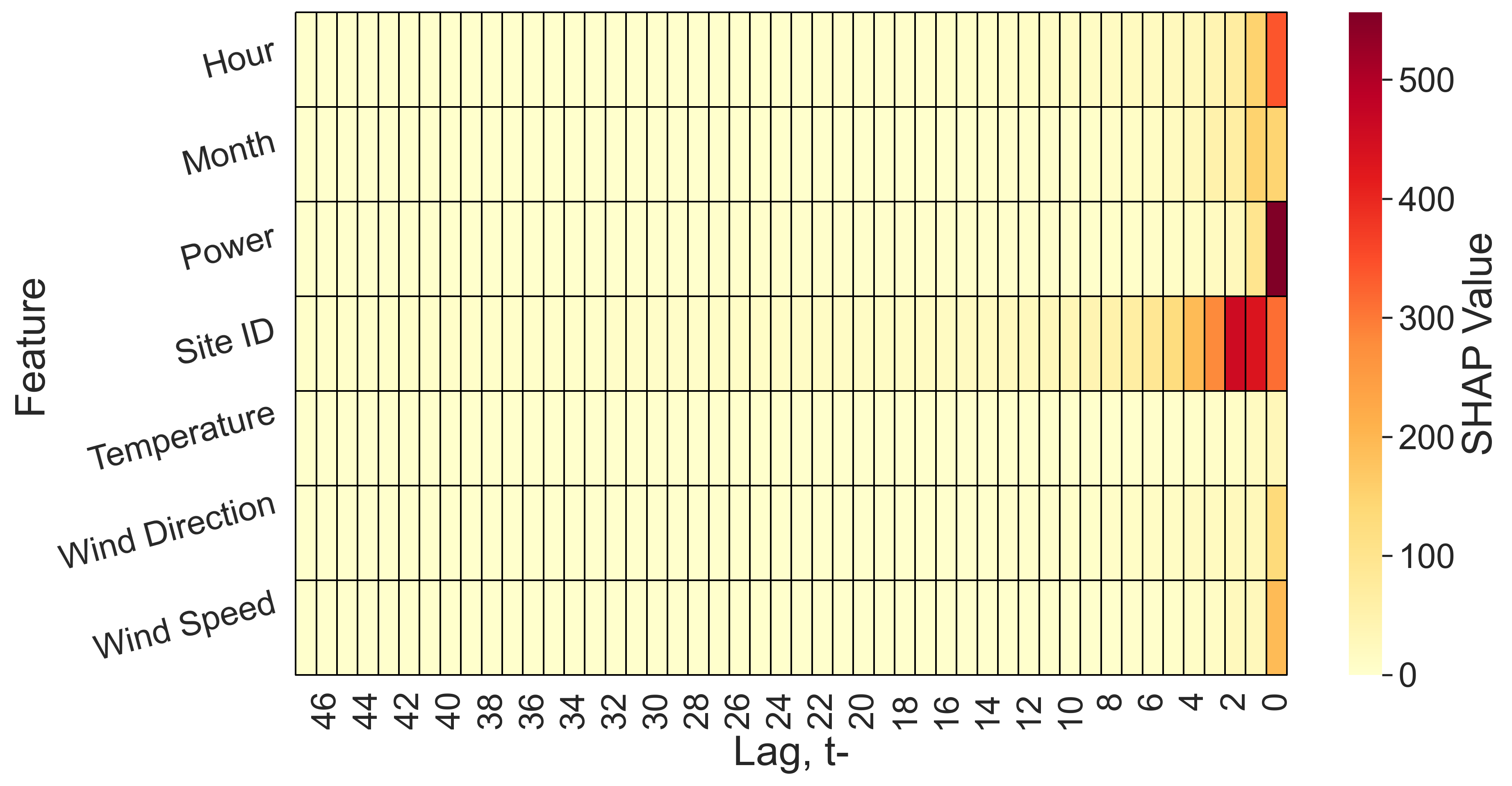}
        \caption{Across Lags}
        \label{fig:heat_features_lags}
    \end{subfigure}
    \begin{subfigure}[t]{.495\textwidth}
        \centering\includegraphics[width=1\textwidth]{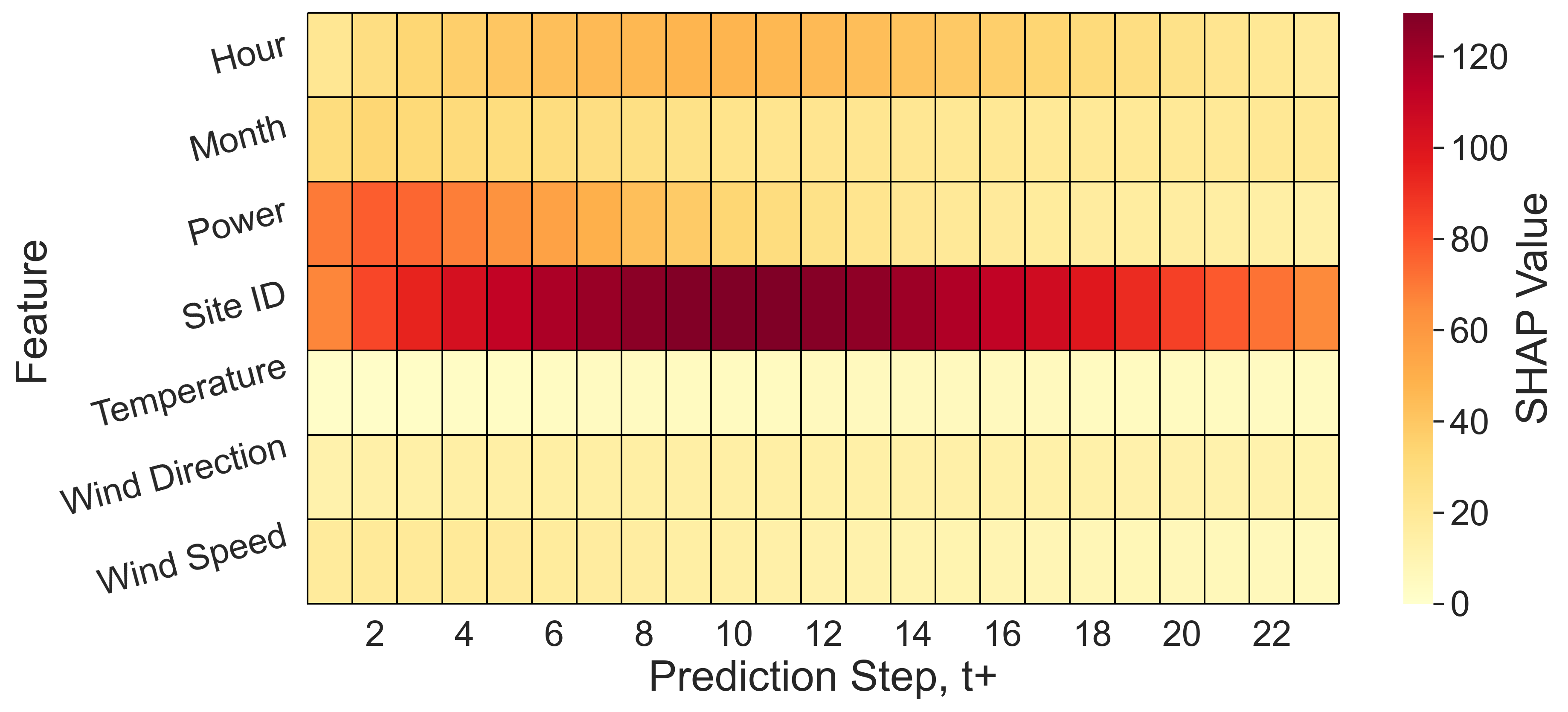}  
        \caption{Across Prediction Steps}
        \label{fig:heat_features_prediction}
    \end{subfigure}
    
  \caption{Heatmaps for feature importance.}
  \label{fig:heatmap_features}
\end{figure*}

The summary plots in Figures \ref{fig:heat_features_lags}, \ref{fig:heat_features_prediction} show the SHAP values across both lags and prediction steps. Here, the absolute SHAP value is used as a summary to show the overall impact. Across all visualisations, the most dominant features are \textit{site ID} and \textit{power}. However, it is clear from figure \ref{fig:heat_features_lags} that most of this attribution for power comes from the most recent time step \textit{t-0} and is only predictive for the first $7$ prediction steps. Beyond this, site ID and therefore site-specific factors captured by \textit{site ID} (and minorly \textit{month}) take over, especially in the mid-range prediction steps. Additionally, \textit{temperature} has negligible importance while the remaining features have a small amount with the most importance attributed to $t-0$.

\begin{figure}
        \centering
        \includegraphics[width=1\linewidth]{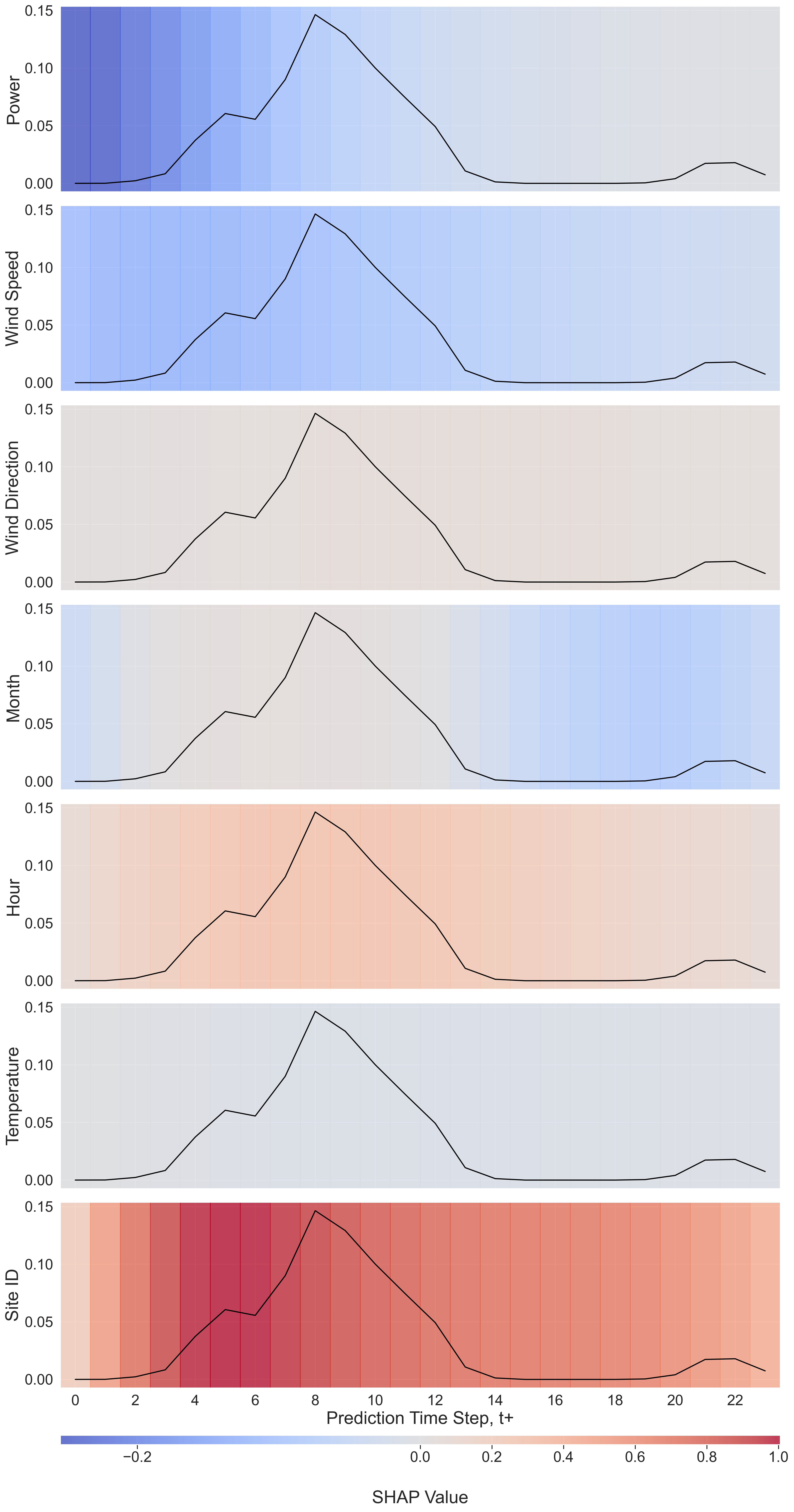}
    
  \caption{Temporal heatmaps for a single samples features across prediction steps.}
  \label{fig:heat_ind}
\end{figure}

In contrast, the heatmaps in Figures \ref{fig:heat_ind} and \ref{fig:heat_ind_hist} provide instance-specific insights and local explanations with directional SHAP values.  For the instance recorded, there is a peak in power production observed. During this peak, \textit{site ID} increases in importance, while \textit{power} decreases. \textit{Hour} shows a slight peak in importance, while \textit{month} drops to zero during the peak. Additionally, positive SHAP values for \textit{site ID} and \textit{hour} indicate their role in predicting higher outputs, while other features, including \textit{power}, predominantly predict lower outputs. While the latter may seem counterintuitive, both observations suggest that more stable periods rely on \textit{power}, \textit{wind speed} and monthly seasonality, while diurnal patterns and site-specific factors inform predictions during periods of instability.

\begin{figure}
    \centering
    \includegraphics[width=.5\textwidth]{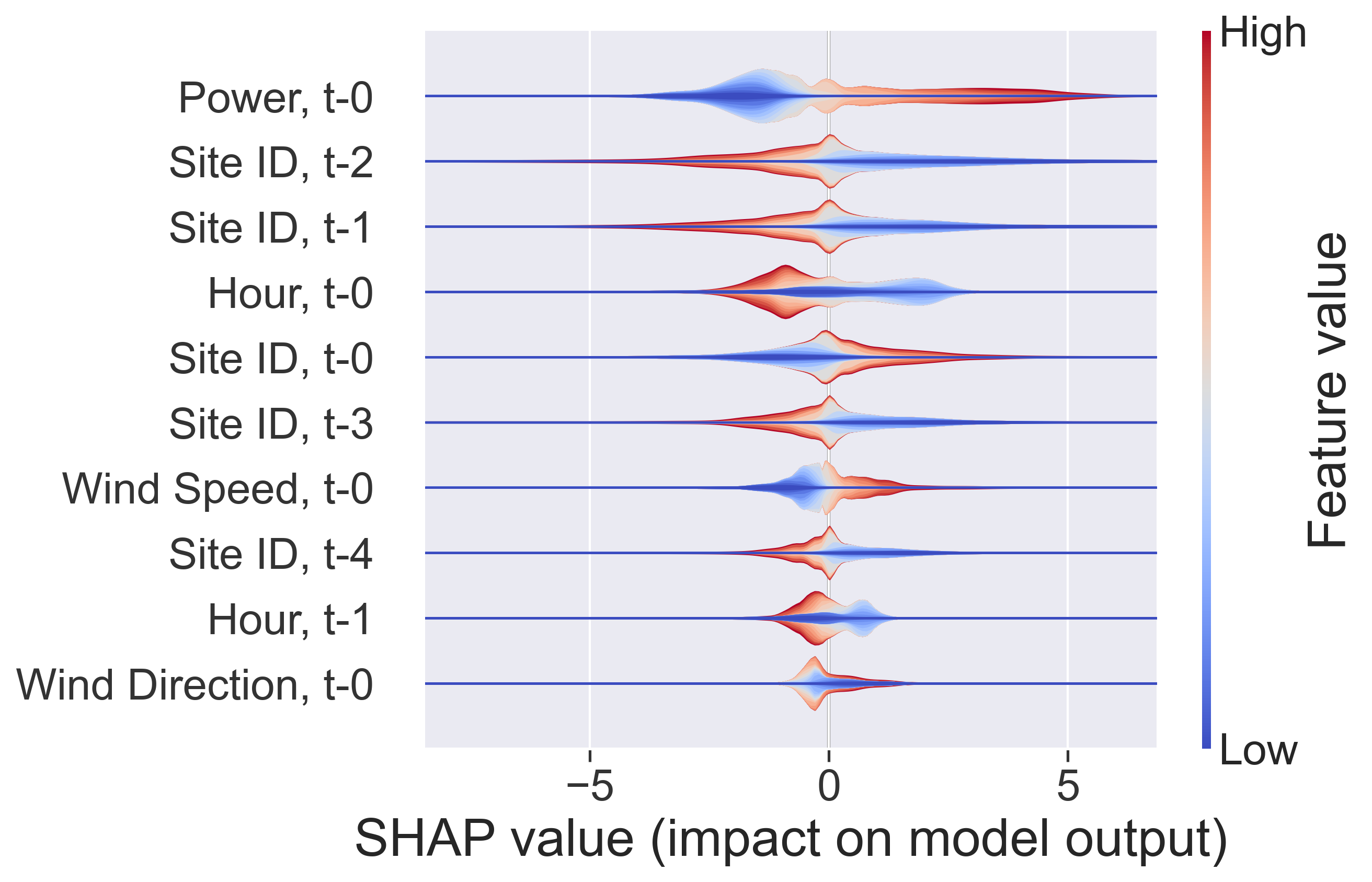}
    \caption{Layered Violin Plot}
    \label{fig:orig_shap}
\end{figure}

Additionally, the layered violin SHAP plot in Figure \ref{fig:orig_shap} shows the top $10$ most important features in order of importance. These plots allow analysis of the most predictive, individual features, their directional impact, feature values, and distributions. The results align with earlier findings, emphasising that \textit{power t-0} is the most important feature, with high feature values leading to positive predictions and low values leading to negative predictions. \textit{Hour} is important at both $t-0$ and $t-1$, with its distribution seeming to align with diurnal patterns in that early hours reduce prediction. Finally, \textit{wind speed} at \textit{t-0} and \textit{wind direction} at \textit{t-0} are also significant, with \textit{wind speed} mirroring the relationship of \textit{power} to its feature values.

\subsection{Analysing Interactions}
\label{sec:interactions}
In this section, we investigate the dominant role of \textit{site ID} to determine if its influence can be attributed to site-level variations in other features. 

To explore potential interactions we use SHAP dependency plots. These plots depict the SHAP values for each \textit{site ID}, coloured by various feature values. However, since \textit{site ID} is a categorical feature, a beeswarm-style plot is more appropriate in this case. This allows for a clearer visualisation of the distribution of SHAP values across all instances, effectively showing the variance and spread of SHAP values at each \textit{site ID}. Furthermore, because of the negligible importance of earlier time steps, we use the $10$ most recent time steps to capture interactions that are most likely to affect prediction outcomes. Moreover, to account for the additional dimension of prediction time steps, the SHAP values are summed across all prediction horizons. This summation offers an overall view of the feature's impact on the model output.

\begin{figure*}[hp!] 
    \centering
    \vspace{-22em} 
    \begin{subfigure}{1\textwidth}
        \centering
        \includegraphics[width=\textwidth]{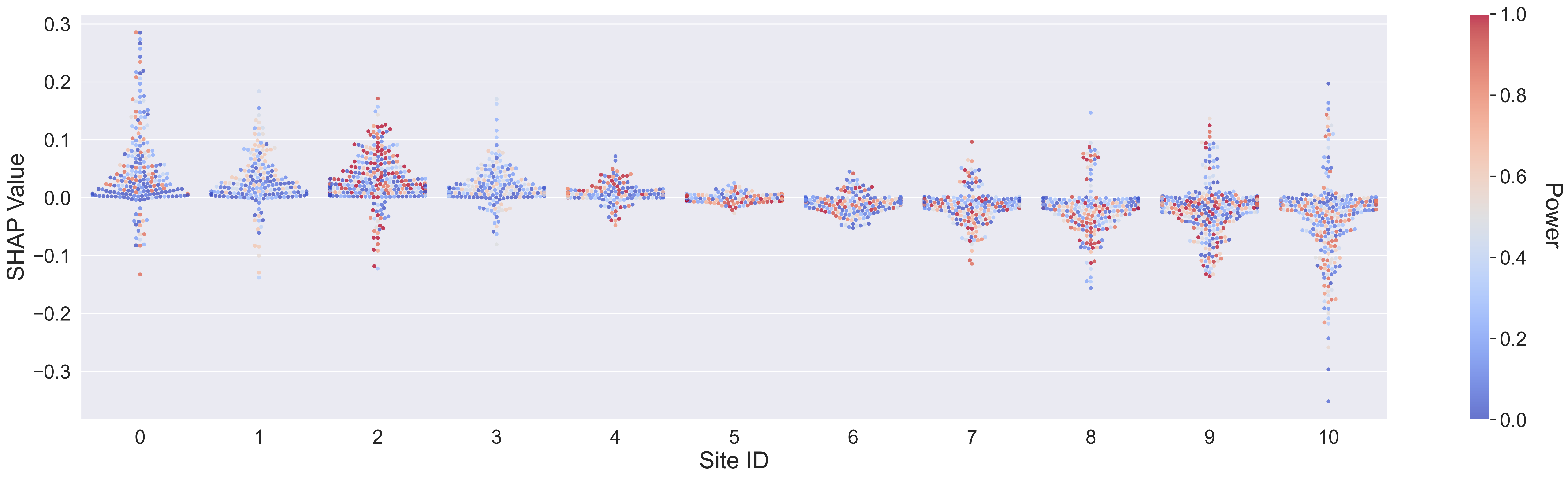}
        \caption{Power}
        \label{fig:shap_dp_power}
    \end{subfigure}

    \begin{subfigure}{1\textwidth}
        \centering
        \includegraphics[width=\textwidth]{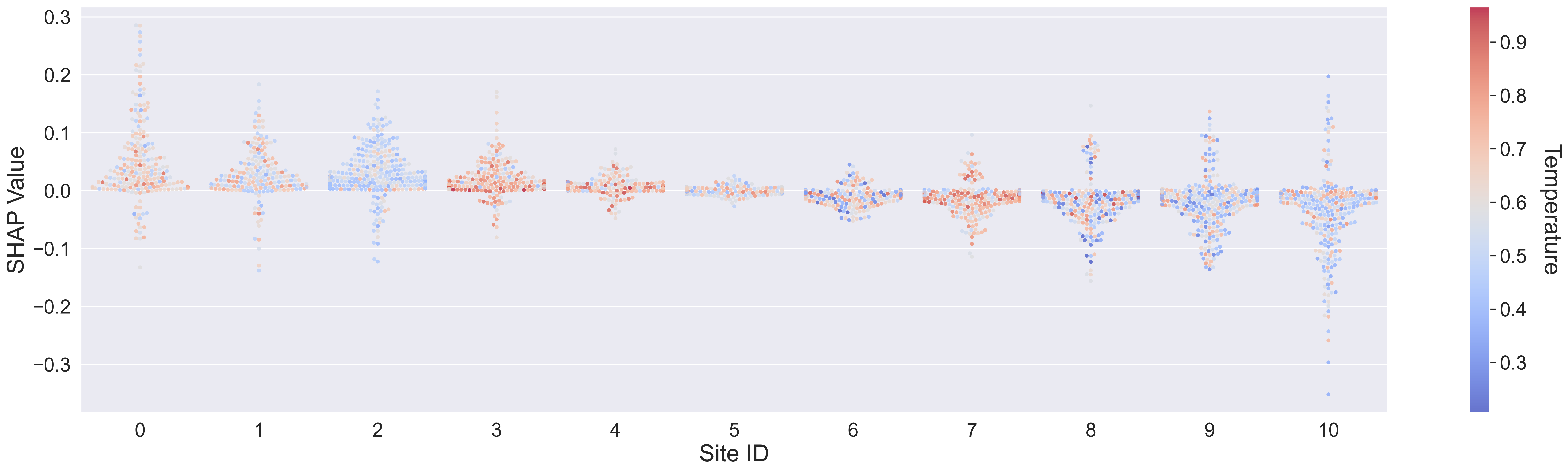}  
        \caption{Temperature}
        \label{fig:shap_dp_temp}
    \end{subfigure}

    \begin{subfigure}{1\textwidth}
        \centering
        \includegraphics[width=\textwidth]{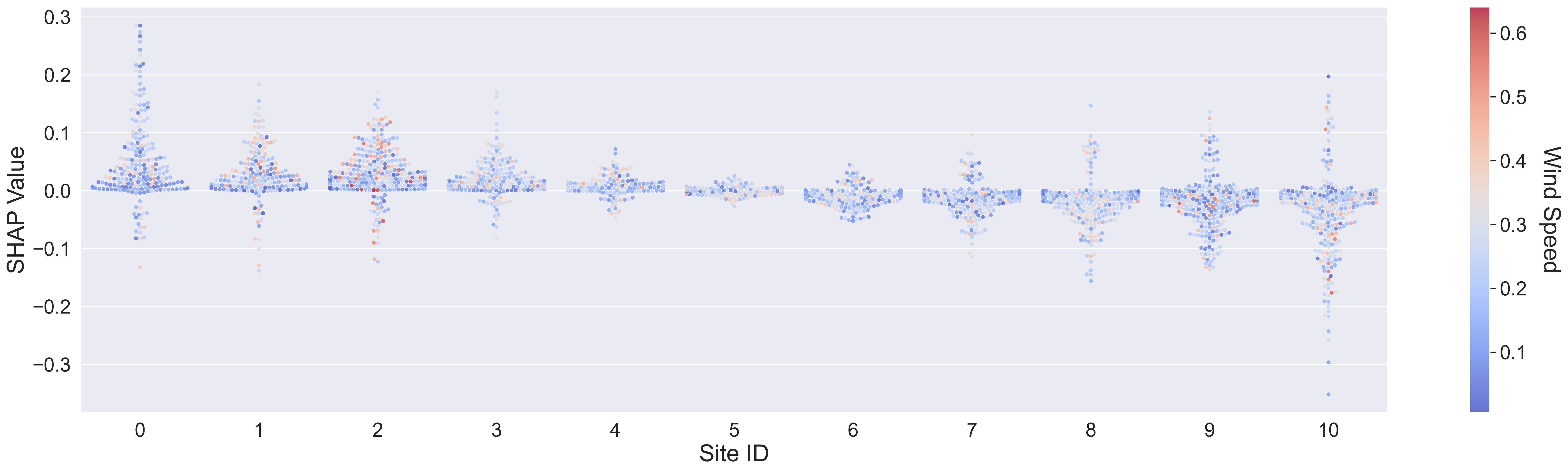}
        \caption{Wind Speed}
        \label{fig:shap_dp_wind_speed}
    \end{subfigure}

    \begin{subfigure}{1\textwidth}
        \centering
        \includegraphics[width=\textwidth]{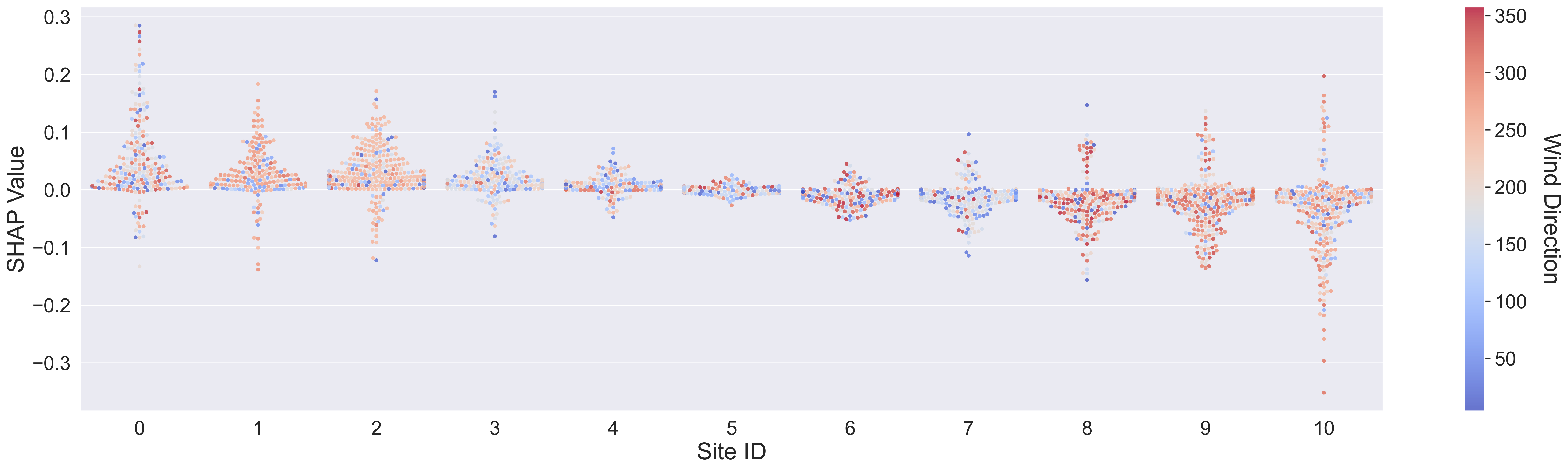}  
        \caption{Wind Direction}
        \label{fig:shap_dp_wind_direction}
    \end{subfigure}

    \caption{SHAP Dependency plot for site ID coloured by all features, using \textit{t-10} to \textit{t-0}. Values have been aggregated over forecast horizons.}
    \label{fig:shap_dp}
    \vspace{-15em} 
\end{figure*}

The dependency plots in Figure \ref{fig:shap_dp} reveal a significant variation in SHAP values between sites. Sites $0$ (Washington) and $10$ (Texas) consistently exhibit higher SHAP values, indicating stronger feature interactions at these locations. In contrast, sites $4$ (New York), $5$, and $6$ (Iowa) show lower SHAP values, suggesting reduced variability or weaker environmental interactions. 

No obvious relationship is seen in Figures \ref{fig:shap_dp_power} and \ref{fig:shap_dp_wind_speed}, however, the higher values of power are not present in all sites, specifically, sites $0$, $1$ and $3$. Despite this, they have some of the highest SHAP values out of all the sites, reiterating what was found in the layered violin plot and suggesting confounding factors. As for the \textit{wind speed} plot, high wind speeds appear to be exceptionally rare and only really present in sites $2$, $9$, and $10$, suggesting that these sites have higher wind variability and overall maximum wind speeds, but this is not directly influencing the model.

Analysis of \textit{temperature} and \textit{wind direction} (Figure \ref{fig:shap_dp_temp} and \ref{fig:shap_dp_wind_direction}) reveals further variation between sites. Generally, higher temperatures seem to occur more centrally, while lower values appear at the extremes in SHAP value. This can mostly be explained by high temperatures generally being associated with atmospheric stability, leading to reduced wind speeds and therefore reducing the influence of \textit{temperature} as it corresponds with lower variability and weaker winds. This also explains why \textit{site ID} becomes less important in sites with a majority of high temperatures, as power generation is more predictable by general trends. Additionally, sites $3$, $4$ $5$ and $7$ have a significantly higher proportion of low values of \textit{wind direction} and higher variability. Interestingly this also aligns with the temperature values observed previously. While there is no clear link overall, the SHAP extremes have matched feature value magnitude for most sites. This suggests that \textit{wind direction} can either enhance or reduce Site ID’s importance, depending on local wind patterns and turbine alignment or yaw control\cite{heMachineLearningbasedFatigue2022}.

Although some useful interpretations can be made from these plots, it can become unclear when comparing the feature values on the same scale. It would be useful to colour the sites by feature value individually to uncover more localised effects that may be masked.

\subsection{Verifying Relationships}

\begin{figure*}[htbp]
\vspace*{-5em}
    \centering
    \begin{tikzpicture}[scale=1, every node/.style={scale=1}]
        
        \node[anchor=north west, inner sep=0, scale=1] (image) at (0, 0) {
         \includegraphics[width=\textwidth]{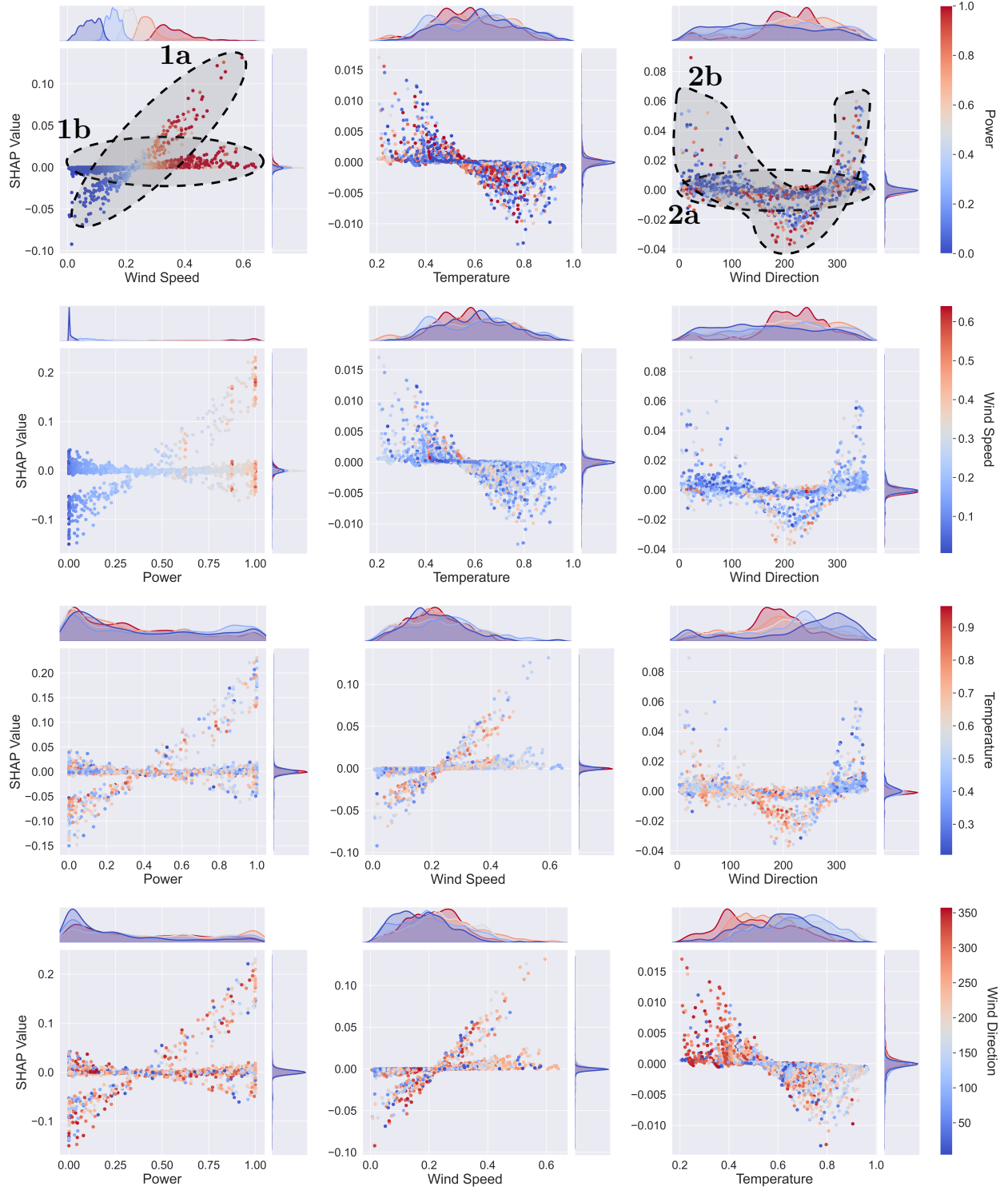}
        };
         \begin{pgflowlevelscope}{\pgftransformscale{1.35714285714}}
        \draw[thick, black, fill=gray, fill opacity=0.2, dashed] (1.95, -2) ellipse (1.2cm and 0.3cm);
        \node[anchor=north east, font=\bfseries] at (2.4, -0.45) {1a};

        \draw[thick, black, fill=gray, fill opacity=0.2, rotate=45, dashed] (0.1, -2.55) ellipse (1.45cm and 0.4cm);
        \node[anchor=north west, font=\bfseries] at (0.5, -1.35) {1b};

        \draw[thick, black, fill=gray, fill opacity=0.2, dashed] (9.4, -2.35) ellipse (1.2cm and 0.25cm);
        \node[anchor=north east, font=\bfseries] at (8.6, -2.4) {2a};
        
        \path[draw=black, fill=gray, fill opacity=0.2, thick, rounded corners, dashed]
        (8.2, -1.05) 
        .. controls (8.1, -2.25) .. (9.1, -2.67)
        
        .. controls (9.3, -3.5) and (10.5, -3.1) .. (10.55, -1.1) 
        --
        (10.1, -1.2) 
        
        .. controls (10.25, -2.9) and (9.25, -2.4) .. (8.8, -1.3) 
        -- cycle; 
        \node[anchor=north west, font=\bfseries] at (8.2, -0.7) {2b};
        \end{pgflowlevelscope}

    \end{tikzpicture}
    \caption{SHAP dependency plots for all features from $t-10$ to $t-0$, with observational cluster annotations. Clusters \textit{1a} and \textit{1b} (\textit{wind speed}) and \textit{2a} and \textit{2b} (\textit{wind direction}) are identified in the top row only (coloured by \textit{power}).}
    \label{fig:known_rels}
\end{figure*}
The SHAP Joint plots (Figure \ref{fig:known_rels}) provide a detailed overview of SHAP dependency plots alongside KDE plots discretised into feature value ranges, for both the $x$-axis and $y$-axis, enabling the identification of high- and low-density regions in the feature space. This allows for visual verification of known relationships in the model while highlighting the impact of secondary features to investigate interactions further. Each row in the figure corresponds to a different primary feature, with each plot coloured by a secondary feature. 

The first row, coloured by \textit{power}, demonstrates a largely expected relationship between \textit{wind speed} and SHAP values (first column): \textit{wind speed} is strongly and positively correlated with the SHAP value and therefore power output. However, two clear clusters form, one which remains centred around a SHAP value of $0$ (\textit{1a}) and one which follows this positive relationship (\textit{1b}). The \textit{wind direction} plot (rightmost column) reveals a potential explanation, forming another two clusters, one centred around zero (\textit{2a}). In contrast to the \textit{wind speed} and \textit{power} plots, there seems to be another cluster following a quadratic relationship with a minimum around $200\circ$ (\textit{2b}). The latter indicates that \textit{wind direction} may be overriding the importance of \textit{wind speed} and \textit{power} in these instances. The cluster \textit{1a} in the \textit{wind speed} and power plots may correspond to points that likely reflect suboptimal wind direction for turbine alignment, suppressing the importance of other features. However, given the presence of cluster \textit{2b}, there are likely other confounders at play, such as wind shear. \textit{Temperature} (second column) shows a negative correlation with the SHAP value. As previously mentioned this is expected, yet, this is invalidated when colouring \textit{wind speed} by \textit{temperature} (third row, second column) as there is no clear correlation. This may be accounted for by the site specific factors or inefficiencies at higher temperatures.

The second row offers further insight now focusing on \textit{wind speed}. Generally, points with low wind speeds adhere to the expected pattern (first column): \textit{power} is strongly and positively correlated with the output power and therefore SHAP value. Furthermore, the same type of clustering appears as in the previous wind speed plot when it comes to SHAP values (\textit{1a}, \textit{1b}), albeit with more spread at the power extremes for cluster \textit{1a}. Most of the wind speed values in the wind direction plot (rightmost column) are evenly distributed, however, higher values seem to cluster around $200\circ$, like the minimum of cluster \textit{2b}. Despite this high wind speed, the ones within this cluster have a low SHAP value. It is unclear from the KDE plot for the $y$-axis as to which cluster these high wind speed values correspond to, but it is clear that the relationship between \textit{wind speed} and output breaks down in cluster \textit{2b}, with the lowest SHAP values having the highest wind speeds and the samples with the highest SHAP values have the lowest wind speeds. Therefore, it may be the case that cluster \textit{2b} corresponds with cluster \textit{1a}.

The \textit{power} plots coloured by \textit{temperature} (third row) reveal further context for this. Within the \textit{wind direction} plot (rightmost column) the highest values are \textit{temperature} are clustered around the minimum of \textit{2b}, $200\circ$, with lower values clustered at larger angles and the KDE shows a negative relationship between the two features. Taking this together, if we assume clusters \textit{2b} and \textit{1a} are one of the same, given that \textit{1b} does not follow the assumed relationship regarding \textit{wind speed}/\textit{power} and SHAP value, then \textit{temperature} may be degrading the attribution of \textit{wind speed} and \textit{power}. This can further be seen in the last plot coloured by \textit{wind speed} (third row, third column) within the distribution of \textit{wind direction} values. This can generally be explained by higher temperatures being associated with atmospheric stability where warmer air creates pressure gradients that drive winds. Conversely, at lower temperature, there may be greater temperature gradients, which create strong winds and variability making \textit{wind direction} more influential. Therefore, there could be a pressure gradient causing a suboptimal change in wind direction and reducing the importance of other features in cluster \textit{1a} which could also explain the correlation found between SHAP value and \textit{temperature}..

Given its influence on pressure gradients and the interactions between \textit{temperature} and \textit{wind direction}, it may have also been beneficial to include atmospheric pressure data for a more complete explanation. Additionally, further insights could have been gained by confirming the clusters and identifying what sites they belong to.

\subsection{Temporal Analysis}
\begin{figure*}[htbp]
    \centering
    \begin{subfigure}[t]{.49\textwidth}
        \centering\includegraphics[width=1\textwidth]{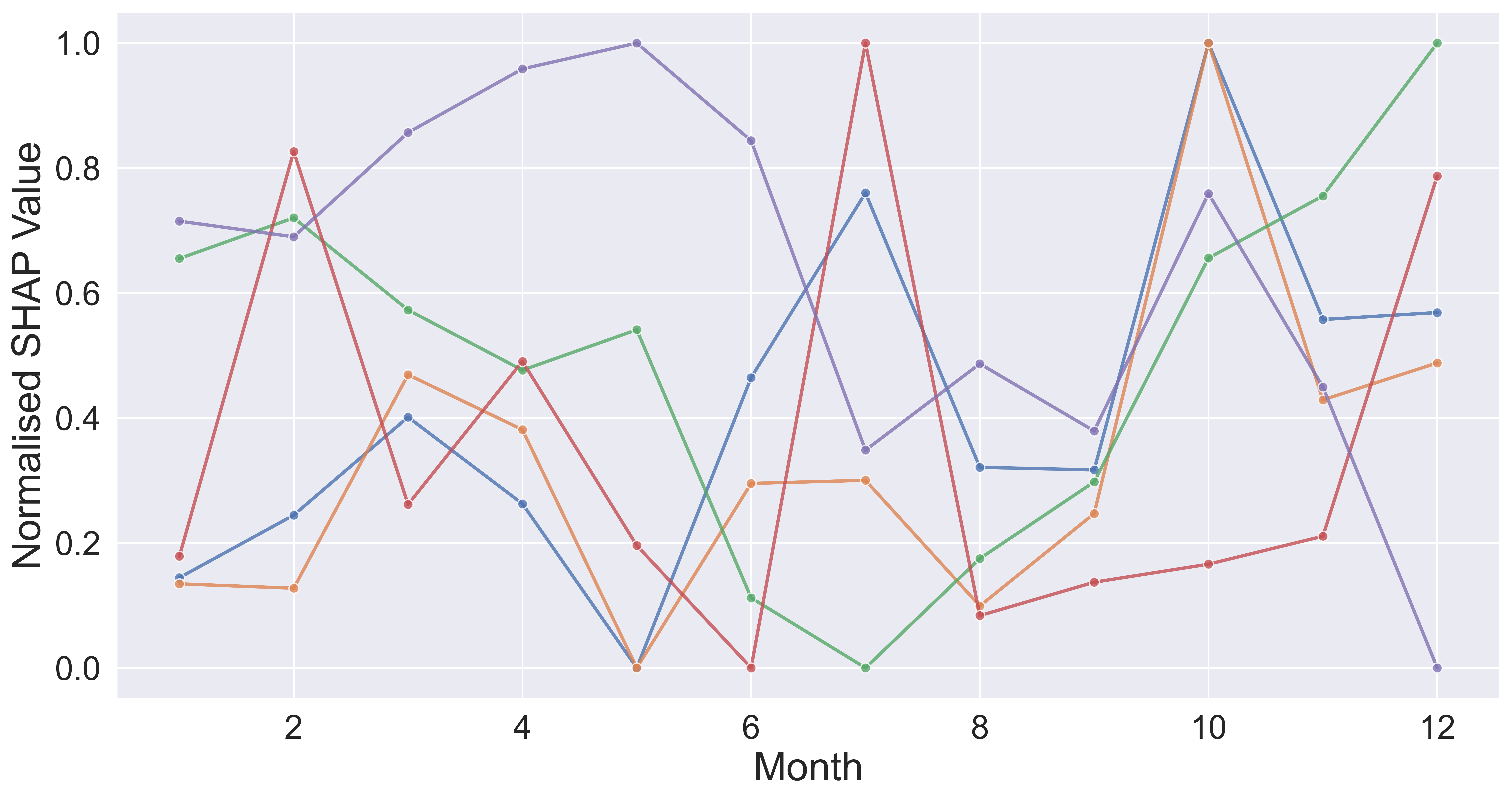}
        \caption{Month}
        \label{fig:shap_hour}
    \end{subfigure}
    \begin{subfigure}[t]{.49\textwidth}
        \centering\includegraphics[width=1\textwidth]{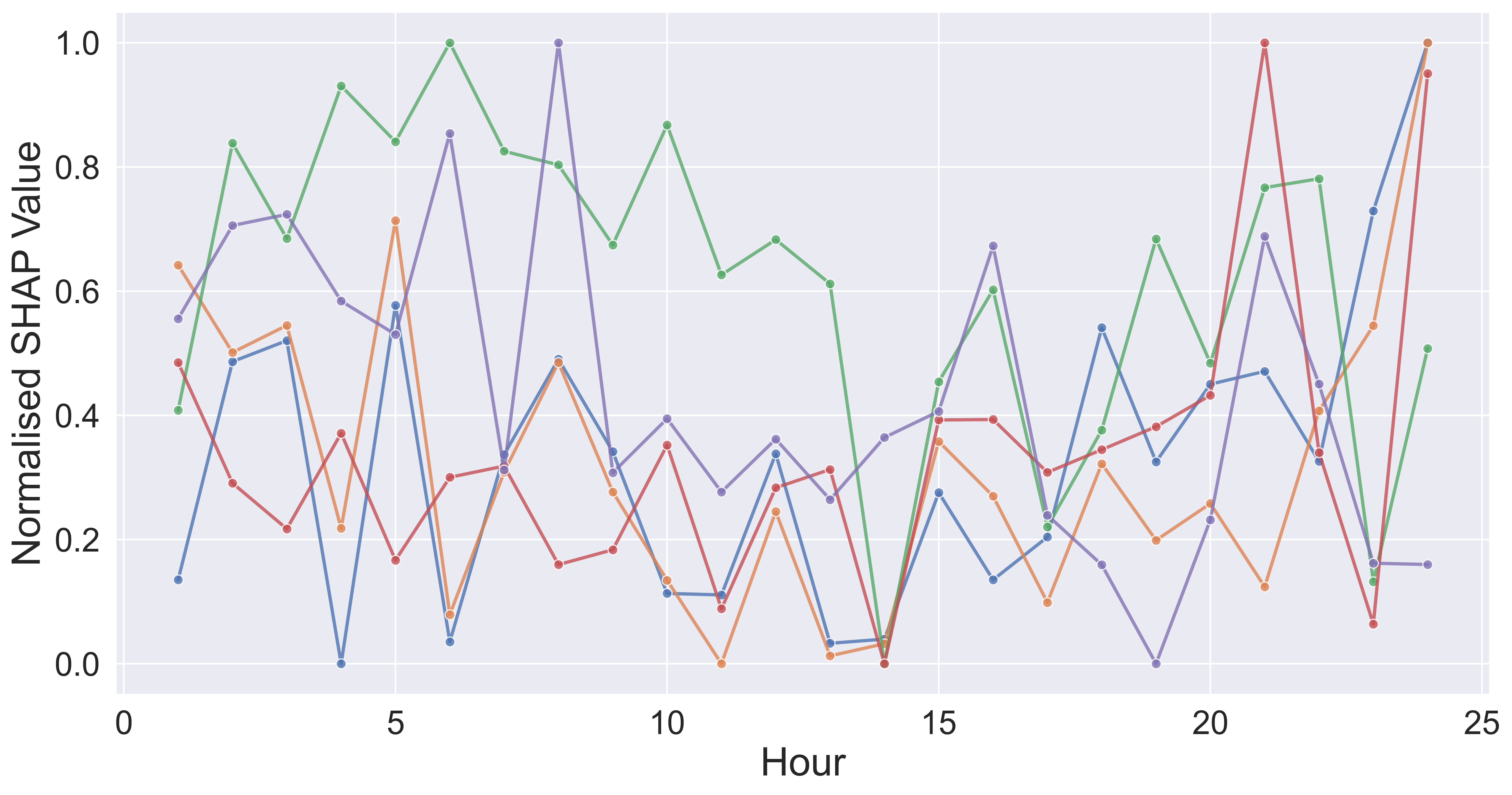}  
        \caption{Hour}
        \label{fig:shap_month}
    \end{subfigure}

    \centering
    \includegraphics[width=.75\textwidth]{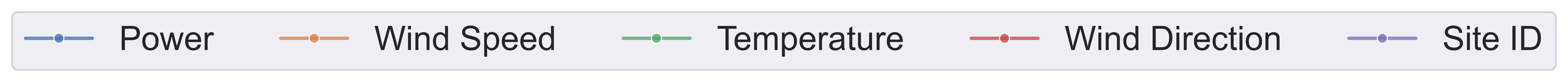}

    \caption{Temporal aggregation of SHAP values at \textit{t-0} by \textit{month} and \textit{hour}.}
    \label{fig:shap_temporal}
\end{figure*}

The temporal SHAP analysis (Figure \ref{fig:shap_temporal}) examines how feature importance changes over time, specifically focusing on seasonal and diurnal patterns. We aggregated SHAP values over the \textit{month} and \textit{day} features to show how feature importance evolves with changing meteorological conditions, similar to \cite{bialekDatasetLevelExplanation2022}. Additionally, we choose to use the final feature for simplicity and normalise the SHAP values across all features for comparability.

In the \textit{month} plot (Figure \ref{fig:shap_month}), we observe that \textit{wind speed} and \textit{power} appear to experience very similar changes across time. Both exhibit a seasonal dip between March and June, which coincides with known seasonal trends. Both \textit{Wind speed} and \textit{temperature} tend to also be lower during the warmer months leading to an expected reduction in SHAP value. Given the prior relationships found earlier, \textit{temperature} also dips in these months, potentially leading to lower SHAP values. Both features also peak in October and to a lesser extent in March. 

Both \textit{power} and \textit{wind direction},  show a peak in July, with the latter a sharply reaching the maximum SHAP value. This can be explained by the prior discussion of \textit{temperature} and atmospheric stability, whereby more organised and consistent patterns can be found in warmer months. In December, both \textit{temperature} and \textit{wind direction} see significant peaks, with the former reaching the maximum. In this month, \textit{power} and \textit{wind speed} are moderate, and \textit{site ID} is at the minimum. \textit{temperature} dominates. This may suggest that \textit{wind direction} becomes more predictable or aligned with turbine placement during colder months, while \textit{temperature} variability has a larger effect on power production.

The final peak within \textit{site ID} can be seen in February, where \textit{site ID} and \textit{temperature} are moderate but \textit{power} and \textit{wind speed} are of low importance. This could indicate site-specific effects, such as geographical location or regional climate conditions that drive differences in turbine performance, even when other meteorological factors are not highly significant. Generally, the importance of \textit{site ID} seems to be reduced when the importance of other features is high and increases when the importance of other features is low, suggesting that during these periods, site-specific factors like terrain or localised wind patterns become more influential. These periods likely correspond to periods of stability and instability and likely reflect the prior findings on local explanations in Section \ref{sec:local} are likely also global patterns. 

\textit{Temperature} shows clear seasonality, peaking in the winter months and gradually decreasing to a minimum in July. This aligns with colder temperatures introducing more variation in turbine operation due to factors like efficiency loss during hotter periods, or more stable wind patterns, demonstrating the model’s sensitivity to temperature effects on turbine performance. Overall, all features show three distinct peaks and troughs, aligning with the seasons or responding to other feature changes affected by the season.

The hourly SHAP analysis (Figure \ref{fig:shap_hour}) offers valuable insights into how feature importance varies throughout the day.  SHAP values for \textit{wind speed} and \textit{power} exhibit notable increases and decreases around midday and late evening, mirroring typical diurnal wind patterns. In addition, most features have a peak around 6 AM and again at 9 PM, with \textit{temperature} being the most notable. These times correspond closely with sunset and sunrise, periods during which temperature gradients caused by solar heating and cooling can lead to shifts in atmospheric stability. The only exception is that \textit{wind speed} dips during the 9 PM peak which could reflect calmer wind conditions due to less thermal gradients at night.

Later in the day, around 2 PM, \textit{temperature} drops to its minimum and is accompanied by a decrease in \textit{wind speed} and \textit{power}. Interestingly, during this same period, \textit{site ID} experiences a slight increase. This pattern may indicate a phase of temperature stabilization, which could affect conditions in the hours that follow.

Additionally, temperature seems to dominate the earlier hours (up to 2 PM), except for two peaks of \textit{site ID}. These observations can again be explained by typical diurnal wind patterns. During the day, solar heating creates turbulence, increasing surface winds, leading to more reliance on \textit{temperature}, while at night, the absence of solar heating leads to a more stable atmosphere with reduced wind speeds. After sunset, cooling reintroduces turbulence, allowing wind speeds to increase again \cite{petenkoTurbulenceLowlevelJets2020, nationalweatherserviceFlightEnvironmentWind}.

\textit{Site ID} tends to have a higher importance in the early hours, dropping around midday before gradually increasing, similarly to \textit{temperature} and \textit{wind speed}, but shows three peaks at 6 AM and 8 AM, where other features (except \textit{temperature}) are substantially lower. This suggests that early morning conditions may depend more on site-specific features, possibly due to terrain effects or microclimates that influence the wind, whereas later in the day, atmospheric stability and general weather patterns take over.

\textit{Power} and \textit{wind speed} seem to peak during the later hours. Specifically, \textit{power} has one main peak at midnight, where \textit{wind speed} and \textit{wind direction} also peak. The others occur at 5 AM and 6 PM, both aligning with \textit{wind speed}. Other than reflecting diurnal variation, these peaks could be caused by increased turbine efficiency when wind speeds are more stable and consistent, either at night or during the early morning and evening transition periods\cite{liuEffectsAtmosphericStability2021}.

Generally, the \textit{hour} plot seems highly sporadic, showing significant changes in importance and power production from hour to hour. Hence, it may have been more informative to sample hourly data from WIND rather than taking the mean. Furthermore, normalising the SHAP values may have been useful to compare relative importance, but may have minimised per-feature trends.

\subsection{Summary}
We conducted a detailed SHAP analysis to interpret the decision-making process of our day-ahead wind power forecasting model, yielding the following key findings:

\begin{enumerate}
    \item \textbf{Site ID's Dominance}: \textit{Site ID} consistently emerges as the most important feature across time steps and prediction horizons. This highlights the significant role of site-specific factors, such as geographical and environmental conditions, which interact with other variables to shape predictions.
    \item \textbf{Importance of Time Steps:} \textit{t-0} consistently dominates feature importance across most variables, with earlier lags contributing negligible influence except \textit{site ID}.
    \item \textbf{Temporal Variability:} Seasonal and diurnal patterns reflect shifts in feature importance. \textit{Power t-0} dominates short-term predictions, while \textit{Site ID} becomes increasingly influential in mid-range predictions. Temperature peaks in importance during colder months, correlating with operational challenges, while diurnal patterns show a relative increase in reliance on site-specific factors during early morning hours.
    \item \textbf{Feature Interactions}: The analysis reveals complex relationships between features. \textit{Wind direction} often overrides \textit{wind speed} when suboptimal alignments or local patterns are present. Clusters in SHAP joint plots confirm that features like \textit{temperature} and \textit{wind direction} interact non-linearly, driven by local conditions and turbine-specific factors.
    \item \textbf{Verified Relationships}: Known relationships, such as the positive correlation between \textit{wind speed} and \textit{power}, are validated. However, clusters of low SHAP values at high wind speeds and suboptimal wind directions suggest suppressed feature importance in certain conditions, likely due to pressure gradients and turbine misalignment.
    \item \textbf{Environmental and Localized Effects:} \textit{Temperature} impacts the model in distinct ways: higher temperatures, associated with atmospheric stability, reduce variability and diminish feature importance, while lower temperatures increase variability in \textit{wind direction} and enhance its predictive role. Site-specific conditions, such as terrain and local wind patterns, also influence model outputs, particularly during transitional periods like early morning and late evening.

\end{enumerate}

Overall, our SHAP analysis largely aligns the model's decision-making process with real-world dynamics while further illuminating the inner workings of the model. By leveraging global and local visualisations, we validated known relationships, uncovered temporal patterns, and identified interactions that inform predictions.

%% file: Sections/Conclusion.tex
\section{Conclusion}
\label{sec:conclusion}

This paper demonstrates the utility of SHAP methods, namely GradientSHAP as a tool for enhancing interpretability in multivariate, multistep time series forecasting. We focus on day-ahead wind power forecasting, a vital task for effective grid management and the global transition toward net-zero energy systems, given the inherent unpredictability of wind power.

Our application of GradientSHAP revealed critical insights into the decision-making process of our forecasting model. \textit{Site ID}, representing site-specific factors such as terrain, environmental conditions, and turbine configurations, emerged as the most important global feature. Its sustained importance across all time steps, particularly in mid-range prediction horizons, highlights its role in capturing contextual and long-term variability, beyond the immediate influence of other features. These findings strongly suggest that \textit{Site ID} should be a core component of forecasting models to capture site-specific variability effectively. \textit{Power t-0}, the most recent historical observation, dominated short-term predictions, reflecting the model's reliance on immediate past data for accurate forecasting. Temporal SHAP heatmaps revealed dynamic shifts in feature importance, with diurnal and seasonal patterns aligning closely with real-world dynamics. These patterns have emphasised the complementary roles of site-specific and meteorological factors. The analysis also highlighted complex interactions between \textit{site ID} and other features, particularly \textit{wind direction} and \textit{temperature}. These findings uncover subtle dependencies, such as the influence of atmospheric stability and pressure gradients, on feature importance.

In addition to uncovering novel insights, the analysis validated known relationships, and interactions, strengthening  confidence in the robustness of the model and its alignment with real-world dynamics.

Despite its strengths, GradientSHAP’s interpretability remains context-dependent, requiring substantial domain expertise to translate SHAP values into valuable insights. Static visualisations provided valuable explanatory power but limited actionable dynamic insights. In addition, out-of-manifold samples and fixed sampling methods may have introduced biases. Integrating on-manifold sampling\cite{fryeShapleyExplainabilityData2020} or counterfactual approaches such as CF-SHAP\cite{albiniCounterfactualShapleyAdditive2022} could improve the explanations.

Future work could improve GradientSHAP’s utility by integrating additional features such as atmospheric pressure to enrich explanations or spatial data, which could further capture site-specific factors. Additionally developing interactive explanation tools that allow dynamic adjustments of lags and prediction steps could provide significantly more explantory power. Furthermore, while our study focused on SHAP values, incorporating analysis of the model's attention weights could complement these insights, allowing richer comparisons and a more comprehensive understanding of the model’s behaviour. Concept discovery, grouping consistently important features into domain-informed clusters, could provide more intuitive and less demanding explanations.

By advancing these methodologies, we can bridge the gap between model complexity and interpretability, paving the way for a wider adoption of AI systems within the energy sector. Ultimately, our findings contribute to a more reliable and transparent approach to energy forecasting, supporting the global transition to sustainable and resilient energy systems.

%% file: Sections/Appendix.tex
\section{Appendix}

\FloatBarrier
\subsection{Supplementary Results}
\begin{figure}[!ht]
    \centering
    \includegraphics[width=\linewidth]{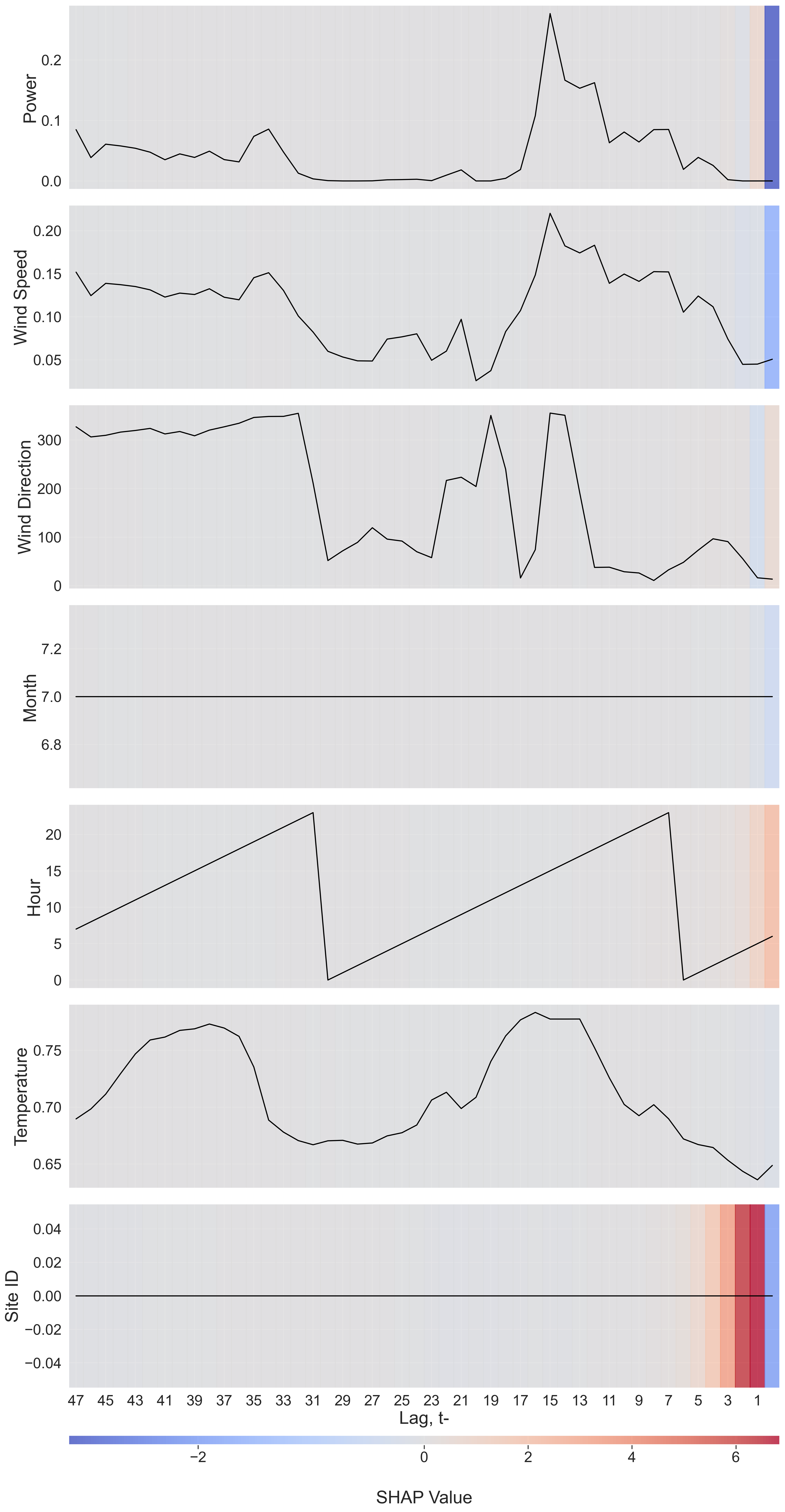}
    \caption{Temporal heatmap for a single sample's features across lags.}
    \label{fig:heat_ind_hist}
       
\end{figure}

\FloatBarrier
\subsection{Model And Optimisation}

\begin{table}[!ht]
\end{table}

\onecolumn
\begin{table}[ht]
\centering
\begin{tabular}{c|c|c|c|c}
    \textbf{Layer} & \textbf{Parameter} & \textbf{Range}  & \textbf{Step} & \textbf{Best Found} \\ \hline
    \multirow{1}{*}{\textit{Embeddings}} & Dimension  & 1 - 20 & 1 & 9 \\
    \hline
    \multirow{4}{*}{\textit{Encoder}} & Dropout rate & 0 - 0.75 & 0.25 & 0.75 \\
    \cline{2-5} & Post layer dropout rate & 0 - 0.75 & 0.25 & 0.75 \\ \cline{2-5} 
    & Num. layers  & 1 - 4 & 1 & 2 \\
    \cline{2-5} 
    & Hidden size  & 4 - 64 & 1 & 30 \\
    \hline
    \multirow{3}{*}{\textit{Decoder}} & Dropout rate & 0 - 0.75 & 0.25 & 0.25 \\
    \cline{2-5} 
    & Num. layers & 1, 4 & 1 & 4\\ \cline{2-5} 
    & Hidden size  & 4 - 64 & 1 & 30 \\
     \cline{2-5} & Attention type & None, general, concat, dot & -& General \\
     \hline
    \multirow{2}{*}{\textit{FC}} & Dropout rate & 0 - 0.75 & 0.25 & 0 \\
    \cline{2-5} 
     & Num. layers & 1 - 3 & 1 & 3 \\ \hline
    \end{tabular}
    \caption{Summary of final optimised model \& training hyperparameters}
    \label{tab:hp_optimization}
\end{table}

\begin{table}[h!]
\centering
\begin{tabular}{l|c|c}
\textbf{Layer (Type)} & \textbf{Output Shape} & \textbf{Param \#} \\ \hline

EncDecAttention & [1, 24] & -- \\ \hline

Embedding (1-1) & [1, 48, 9] & 99 \\ \hline

Encoder (1-2) & [1, 48, 30] & -- \\
\hspace{1 cm} LSTM (2-1) & [1, 48, 30] & 13,440 \\ \hline

Dropout (1-3) & [1, 48, 30] & --\\ \hline

Decoder (1-4) & [1, 1, 30] & -- \\
\hspace{1 cm} LSTM (2-2) & [1, 1, 30] & 29,760 \\ \hline

Attn (1-5) & [48, 1, 1] & -- \\
\hspace{1 cm} Linear (2-3) & [1, 48, 30] & 930 \\ \hline

Linear (1-6) & [1, 30] & 1,830 \\ \hline

\end{tabular}
\medbreak

\centering
\textbf{Total params:} 46,090 \\
\textbf{Trainable params:} 46,090 \\
\textbf{Non-trainable params:} 0 \\
\textbf{Total mult-adds (M):} 1.43 \\
\bigbreak

\textbf{Input size (MB):} 0.00 \\
\textbf{Forward/backward pass size (MB):} 0.30 \\
\textbf{Params size (MB):} 0.18 \\
\textbf{Estimated Total Size (MB):} 0.49 \\
\caption{Summary of final forecasting model architecture.}
\label{fig:summary}
\end{table}

\begin{table}[H]
\centering
\begin{tabular}{c|c|c}
    \textbf{Category} & \textbf{Parameter} & \textbf{Value} \\ \hline
    
    \multirow{2}{*}{\textit{Early Stopping}} 
    & Threshold & 0.0001 \\ \cline{2-3} 
    & Patience & 5 \\ \hline
    
    \multirow{4}{*}{\textit{Learning Rate}} 
    & Initial LR & 0.0001 \\ \cline{2-3} 
    & Scheduler type & Reduce on plateau \\ \cline{2-3} 
    & Factor & 0.1 \\ \cline{2-3} 
    & Patience & 10 \\ \cline{2-3} 
    & Threshold & 0.001 \\ \cline{2-3} 
    & Min. LR & $1 \times 10^{-8}$ \\ \hline
 \textit{Misc} & Batch size & 32\\
    
\end{tabular}
\caption{Summary of fixed hyperparameters. LR = learning rate.}
\label{tab:hps}
\end{table}

\subsection{Preprocessing}
\begin{figure*}[htbp!]
    \centering
    \includegraphics[width=1\textwidth]{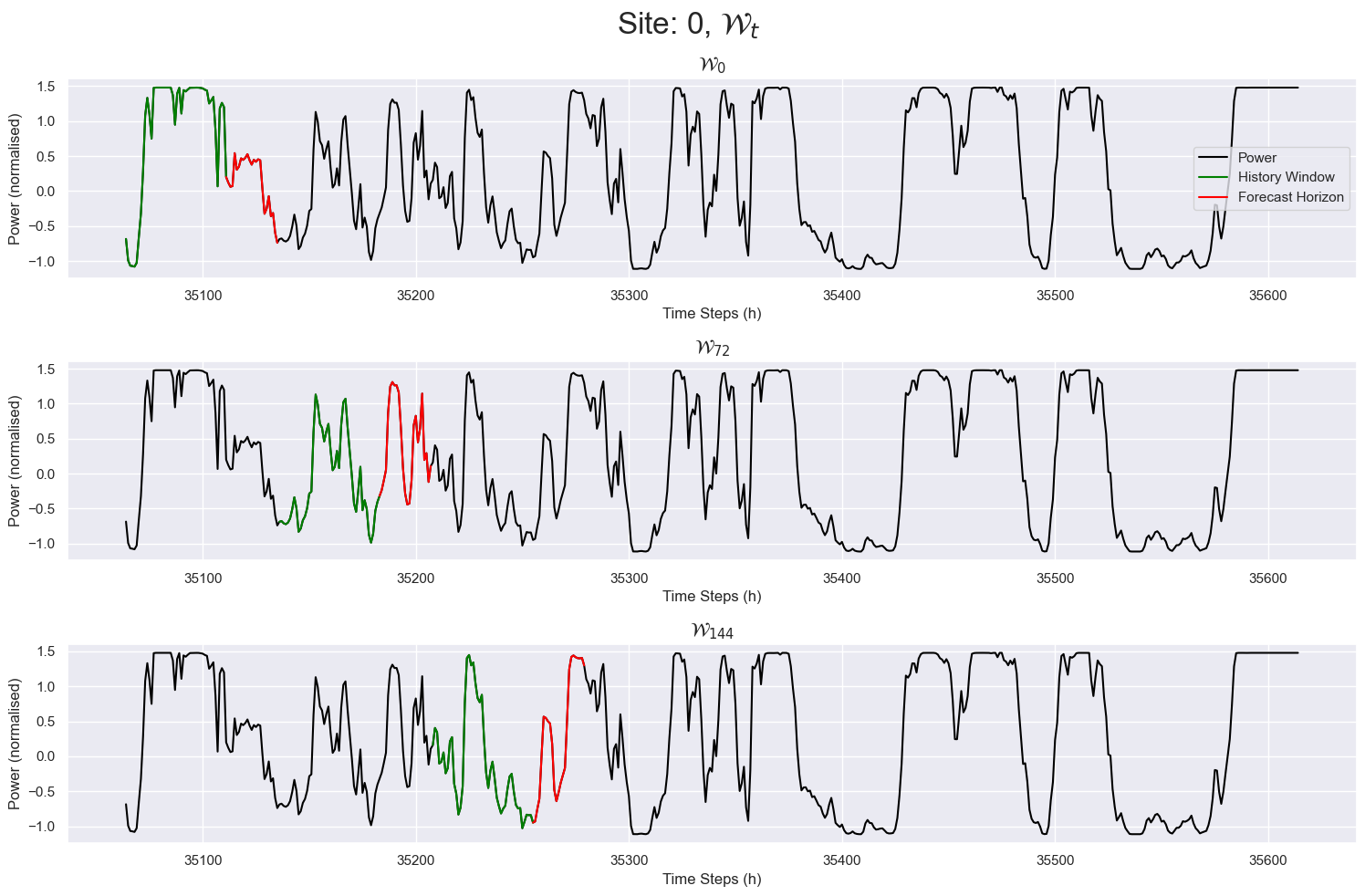}
    \caption{Illustration of three non-overlapping windows for site $0$.}
    \label{fig:rollingwindow}
\end{figure*}

\twocolumn